\documentclass[lettersize,journal]{IEEEtran}

\usepackage{amsmath,amsfonts}
\usepackage{algorithmic}
\usepackage{algorithm}
\usepackage{array}
\usepackage[caption=false,font=normalsize,labelfont=sf,textfont=sf]{subfig}
\usepackage{textcomp}
\usepackage{stfloats}
\usepackage{url}
\usepackage{verbatim}
\usepackage{graphicx}
\usepackage{cite}
\usepackage{amsthm}
\usepackage{amsmath}
\usepackage{amssymb}
\usepackage{mathrsfs}
\usepackage{makecell}
\usepackage{graphicx}
\usepackage{xcolor}
\usepackage[switch]{lineno}

\begin{document}
\title{S2ML: Spatio-Spectral Mutual Learning for Depth Completion}

\author{Zihui Zhao, Yifei Zhang,
Zheng Wang,~\IEEEmembership{Senior Member~IEEE}, Yang Li,~\IEEEmembership{Member~IEEE}, 
Kui Jiang,~\IEEEmembership{Member~IEEE}, \\
Zihan Geng,~\IEEEmembership{Member~IEEE}, 
Chia-Wen Lin,~\IEEEmembership{Fellow~IEEE}

\thanks{This work was supported in part by the National Natural Science Foundation of China (62305184); National Natural Science Foundation of China (62371270); Basic and Applied Basic Research Foundation of Guangdong Province (2023A1515012932); Science, Technology and Innovation Commission of Shenzhen Municipality (JCYJ20241202123919027); Science, Technology and Innovation Commission of Shenzhen Municipality (WDZC20220818100259004). (\textit{Corresponding authors: Zihan Geng; Kui Jiang.})}

\thanks{Z. Zhao, Y. Zhang, Yang Li and Z. Geng are with the Institute of Data and Information, Tsinghua Shenzhen International Graduate School, Tsinghua University, Shenzhen, Guangdong, China, 518071 (emails: zzh23@mails.tsinghua.edu.cn, zhang-yf22@mails.tsinghua.edu.cn, yangli@sz.tsinghua.edu.cn, geng.zihan@sz.tsinghua.edu.cn).}
\thanks{Z. Wang is with School of Computer Science, Wuhan University, 430072, China (email: wangzwhu@whu.edu.cn).}
\thanks{K. Jiang is with School of Computer Science and Technology, Harbin Institute of Technology, 150001, China (email: kuijiang\_1994@163.com).}
\thanks{C.-W. Lin is with Department of Electrical Engineering and the Institute of Communications Engineering, National Tsing Hua University (email:  cwlin@ee.nthu.edu.tw).}
}

\markboth{Journal of \LaTeX\ Class Files,~Vol.~14, No.~8, August~2021}%
{Shell \MakeLowercase{\textit{et al.}}: A Sample Article Using IEEEtran.cls for IEEE Journals}


\maketitle
\begin{abstract}
The raw depth images captured by RGB-D cameras using Time-of-Flight (TOF) or structured light often suffer from incomplete depth values due to weak reflections, boundary shadows, and artifacts, which limit their applications in downstream vision tasks. Existing methods address this problem through depth completion in the image domain, but they overlook the physical characteristics of raw depth images. 
It has been observed that the presence of invalid depth areas alters the frequency distribution pattern. In this work, we propose a Spatio-Spectral Mutual Learning framework (S2ML) to harmonize the advantages of both spatial and frequency domains for depth completion. 
Specifically, we consider the distinct properties of amplitude and phase spectra and devise a dedicated spectral fusion module. Meanwhile, the local and global correlations between spatial-domain and frequency-domain features are calculated in a unified embedding space. The gradual mutual representation and refinement encourage the network to fully explore complementary physical characteristics and priors for more accurate depth completion. 
Extensive experiments demonstrate the effectiveness of our proposed S2ML method, outperforming the state-of-the-art method CFormer by 0.828 dB and 0.834 dB on the NYU-Depth V2 and SUN RGB-D datasets, respectively. 
\end{abstract}

\begin{IEEEkeywords}
Depth completion, RGB-Depth fusion, Fourier domain, Indoor scene
\end{IEEEkeywords}

\section{Introduction}
\IEEEPARstart{D}{epth} sensing is essential for various 3D tasks such as autonomous driving~\cite{chen_deepdriving_2015}, robot navigation~\cite{sabe_obstacle_2004,mahler_learning_2019}, and scene reconstruction~\cite{bascle_stereo_1993,wong_rigidfusion_2021}. However, raw depth images captured by current depth sensors, such as Time-of-Flight (TOF) and structured light devices like Microsoft Kinect~\cite{kinect} and Intel Realsense~\cite{keselman_intelr_2017}, often contain significant invalid areas. These regions arise from factors such as highly reflective or transparent surfaces and challenging lighting conditions. This invalid data hinders the performance of downstream vision tasks that rely on depth images, thereby limiting their practical applications.

To address this issue,  researchers have proposed methods to recover complete depth maps from raw depth images guided by aligned RGB images, a process known as \textit{depth completion}. 
It is challenging to recover fine-grained details across the entire scene by mining correlations between depth and RGB modalities. Although RGB images can provide complementary semantics for invalid depth areas, previous  efforts~\cite{gur2019single,bello2021self,qiao2021vip} indicate that these images often contain inherent depth ambiguity and produce structureless depth features. Thus, a key objective in depth completion is to effectively utilize the complementary information provided by RGB images. 
Some existing methods leverage the affinities of neighboring pixels to refine depth iteratively, framing the task as a diffusion problem~\cite{cheng_cspn_2019,park_non-local_2020,cheng_learning_2019,zhou2022pgdenet,zuo2020frequency}. However, their performance diminishes when faced with large invalid areas lacking neighboring points. To better explore the correlations between depth and RGB modalities, other approaches propose end-to-end designs for feature extraction and depth image restoration~\cite{chen_agg-net_nodate,rho_guideformer_2022,li2022self}. These methods perform depth completion or refinement directly, without an iterative process. Additionally, some efforts~\cite{zhang_completionformer_2023} combine both strategies, applying end-to-end networks for depth completion followed by pretrained iterative models for further refinement. All these works aim to leverage complementary information from the RGB modality to facilitate depth completion in invalid areas.

Despite significant advancements made by these spatial domain methods, many do not fully account for the physical characteristics of invalid areas. These areas often appear in the smooth regions of the original depth image, introducing sharp edges and exhibiting similar properties in the Fourier domain. Neglecting these physical priors can have two potential drawbacks. 
First, the boundary between valid and invalid depth measurements is ambiguous~\cite{liu2021learning}, since the invalid areas in depth maps are typically represented by zero values. Second, the prior knowledge regarding lost information in invalid depth areas mainly consists of texture information, and the resulting low-frequency degradation is frequently overlooked. These issues limit the performance of current multi-modal fusion algorithms in depth completion tasks.

To address these challenges, we propose the Spatio-Spectral Mutual Learning framework (S2ML) for depth completion. This framework gradually integrates information from depth and RGB modalities through a spatio-spectral mutual learning approach. By analyzing the impact of invalid depth areas in the frequency domain, we found that these areas primarily cause specific degradation patterns in different frequency components of the amplitude spectrum and alter the semantic content of the phase spectrum. With this insight, our S2ML method can effectively utilize auxiliary RGB information from both the spectral and spatial domains.

In our spectral fusion module, we apply distinct fusion strategies for the phase and amplitude spectra, considering their unique properties and modes of degradation. The embedded feature maps are transformed into the frequency domain and decomposed into phase and amplitude spectra for specialized fusion. For the phase spectrum, we integrate semantic information from the RGB phase spectrum by extracting spatial-wise features and applying pixel-to-pixel fusion. For the amplitude spectrum, we observe the degradation in various frequency components caused by invalid depth areas and rescale the low-frequency part for band-wise fusion.

In the spatial domain, we integrate spatial features and fused spectral features by leveraging their local and global correlations. Since the depth map encompasses global correlations and requires local precision, we introduce a Swin-Convolution module to achieve feature extraction and correlation perception at both local and global levels. In this module, features from the frequency fusion module are combined with depth spatial features within a unified embedding space.

By employing gradual mutual representation and refinement in our framework, we achieve a comprehensive exploration of complementary physical characteristics and priors. Frequency and spatial domain fusion are conducted recursively to progressively refine depth features and enhance depth completion accuracy. Ultimately, high-dimensional features representing information from both depth and RGB modalities are transformed into depth predictions.

Experiments conducted on the NYU-depth-v2 and SUN RGB-D datasets with different sensors demonstrate the effectiveness and robustness of our proposed method, which significantly outperforms other state-of-the-art methods. To validate our design in Fourier domain fusion, we also evaluate the precise impact of the spectral fusion module on the depth map. Our major contributions are summarized as follows:

\begin{itemize}
    \item We propose a Spatio-Spectral Mutual Learning framework, informed by our analysis of the frequency-domain characteristics of raw depth images, to achieve high-quality depth completion.
    
    \item We introduce a spectral fusion strategy that leverages the distinct properties of both amplitude and phase spectra, effectively utilizing complementary spectral information.
    
    \item We present a spatial domain fusion module that incorporates cascaded Swin-Convolutional modules, designed to extract both global and local information for refining the depth map.
    
    \item Experimental results demonstrate that our proposed method significantly outperforms other baselines on two widely used benchmarks: NYU-Depth v2 and SUN RGB-D.
\end{itemize}

\section{Related works}

In this section, we review previous works on depth completion using multi-modality fusion strategies. Some methods predict dense depth maps by iteratively refining raw depth maps, while others generate refined depth maps directly through an end-to-end approach. Additionally, we examine methods incorporating Fourier-domain information to enhance image restoration and other computer vision tasks.

\textbf{Depth Completion.} The depth completion task aims at constructing a dense depth map from the incomplete depth map. Sparse-to-dense~\cite{ma_sparse--dense_2018} introduced a basic CNN encoder-decoder network to generate dense depth maps. S2dnet~\cite{germain_s2dnet_2020} concatenated RGB and depth images and fed them into a U-Net. Dfusnet~\cite{shivakumar_dfusenet_2019} extracted contextual information from the intensity image and depth map separately and fused them in the subsequent network, which exploits the potential correlations and complementary information within the two modalities. 

Following the spatial propagation network (SPN)~\cite{liu_learning_2017}, SPN-based networks first estimate a rough depth map and iteratively adjust the depth value of each unit. Among them,~\cite{cheng_learning_2019} proposed the convolutional spatial propagation network (CSPN) and introduced long-range context information by a recurrent operation. NLSPN ~\cite{park_non-local_2020} improved CSPN by adopting non-local spatial and global propagation. DySPN~\cite{lin_dynamic_2022}, GraphCSPN~\cite{avidan_graphcspn_2022}, LRRU~\cite{wang_lrru_2023} improved the performance of SPN by further adjusting the paradigm of propagation. 

Apart from these methods,  several end-to-end networks have been proposed to enhance multi-modality fusion~\cite{hu_penet_2021,liu_fcfr-net_2021,zhang2023coarse,xian2023vita,shin2023mosaicmvs,zhao2021depth,ezhov2024all}. A common approach to fusing features from different modalities is through direct concatenation or pixel-wise summation operations. Additionally, more sophisticated fusion strategies have been developed, including channel-wise canonical correlation analysis~\cite{zhong_deep_nodate}, image-guided spatially-variant convolution~\cite{yan_rignet_2022} and attention-based graph propagation~\cite{ma_sparse--dense_2018,zhao_adaptive_2021}.~\cite{rho_guideformer_2022} proposed a token fusion method based on a guided-attention mechanism.~\cite{gao_spatial_2021} adopted a spatial cross-attention module to extract weighted features from the depth branch and apply them to the RGB feature map.~\cite{ezhov2024all} used the abundantly available synthetic data to approximate the 3D scene structure by learning a mapping from sparse to (coarse) dense depth maps along with their predictive uncertainty. These methods are designed to enhance local information interaction and fusion. 

However, these works are primarily conducted in the spatial domain for depth completion, overlooking the physical priors in the frequency domain. In contrast, we propose a mutual learning framework that incorporates frequency-domain information to enhance depth completion performance.

\textbf{Frequency Domain Processing.} Fourier analysis has been extensively applied in the fields of signal processing and computer vision. By utilizing the Fourier Transform to process the input, models acquire the capability for global degradation perception, which serves as the fundamental motivation behind our frequency-domain fusion design. Recently, numerous studies have drawn inspiration from Fourier analysis, incorporating frequency information into neural network designs to fully exploit the potential information contained in input images~\cite{wang2023s,lin2020spatial,peng2022ssml,chen2024frequency}.~\cite{chi_fast_2020} proposed fast Fourier convolution (FFC) to enlarge the receptive field and achieve cross-scale fusion.~\cite{chen_progressively_2020} proposed a frequency method for network pruning by analyzing the redundancy of filters in the Fourier domain.~\cite{wang_high_2020} generalized the convolutional neural network by the spectrum information of the image.~\cite{lin2020spatial} proposed a hyperspectral image classification method based on spectral and spatial information. For objective function,~\cite{jiang_focal_2021} proposed a focal frequency loss to narrow the gap in the Fourier domain between the real image and the predicted image.

As to specific applications,~\cite{zhou_pan-sharpening_2022} integrated spatial-domain and frequency-domain information for multi-spectrum pan-sharpening.~\cite{peng2022ssml} proposed a deep mutual-learning-based framework for spectral-spatial information mining and hyperspectral pansharpening which learns spatial and spectral features from each other through information transmission.~\cite{wang2023s} proposed a dual-branch co-teaching framework by leveraging the intrinsic complementarity of features extracted from the spatial and spectral domains and encouraging cross-space consistency through collaborative optimization.~\cite{zhou2024general} presented a SFINet that integrates spatial and frequency domain information in three different branches which captures both local spatial details and global frequency information, leading to superior performance in pan-sharpening and depth super-resolution tasks.~\cite{lee_single-image_nodate} aggregated estimation candidates in the frequency domain for single image depth estimation.~\cite{mao_intriguing_2023} proposed a frequency-selective network that learns kernel-level features adaptively for image deblurring.~\cite{wang_ffnet_2022} introduced a frequency fusion network that exploits frequency correlation between the depth feature and RGB feature for semantic information. These studies have demonstrated the potential of incorporating frequency-domain information for computer vision tasks. However, they have not fully exploited the distinct characteristics of the amplitude and phase spectra. For image restoration tasks (e.g., deraining, deblurring, pan-sharpening), degradation primarily affects the amplitude spectrum since the semantics remain largely unchanged. In the depth completion task, the semantics of raw depth images change due to information loss in invalid depth areas. In our method, we take into account the physical properties of both the amplitude and phase spectra and propose a frequency fusion module accordingly.

\section{Method}

In this section, we introduce our proposed Spatio-Spectral Mutual Learning framework (S2ML). To effectively leverage the complementary information from both the depth and RGB modalities and the physical priors of raw depth images, we extract and integrate information from both the frequency and spatial domains through progressive representation learning and refinement.

\subsection{Problem Formulation}

Given aligned raw depth image $D^\mathrm{raw}\in{H\times W}$ and RGB image $I\in{H\times W\times 3}$, an image-guided depth completion model $f_{\theta}(R, I)$ aims to recover the invalid areas in the raw depth image and predict a complete depth map $D^\mathrm{pred}$ for downstream computer vision tasks. The RGB image $I$ provides rich semantic information, serving as a complementary modality to the depth image.

For raw depth image $D^\mathrm{raw}$, invalid depth regions are usually caused by factors including highly reflective or transparent/semi-transparent surfaces and lighting conditions. In these regions, the measured depth values drop to zero, creating sharp edges between valid and invalid depth areas. Consequently, for an entire region of invalid depth data, the pixels are typically situated on the same plane, and the corresponding depth values exhibit gradual and uniform changes.

\begin{figure}[t]
	\centering
	\includegraphics[width=\linewidth,scale=1.00]{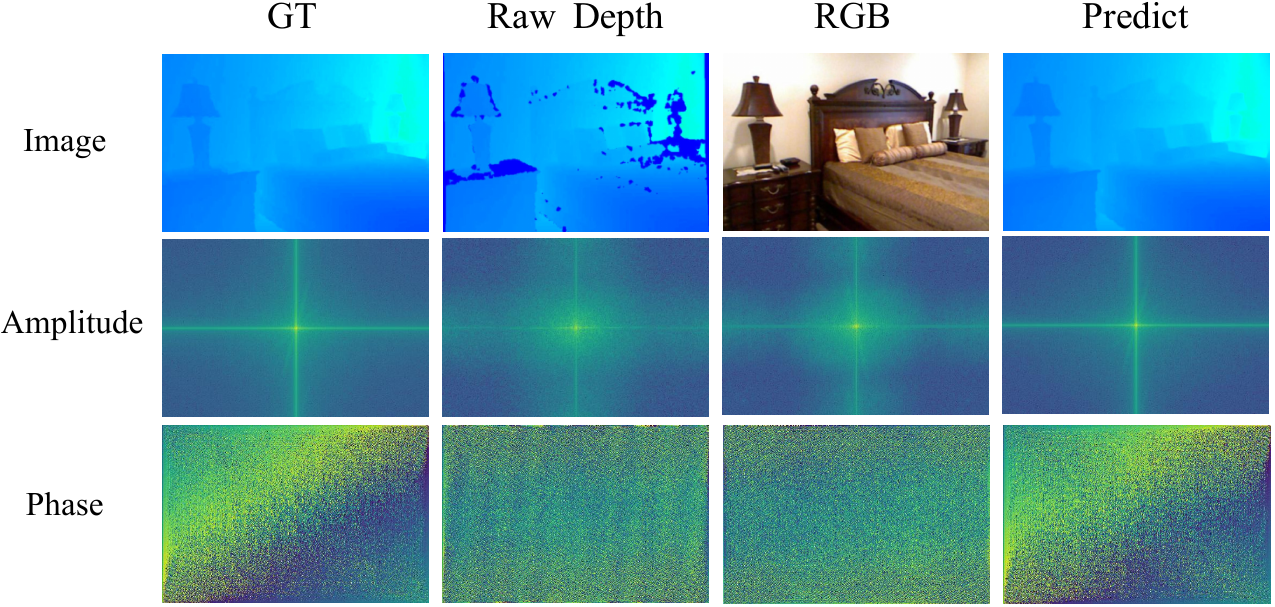}
	\caption{Phase and amplitude spectra of the ground truth depth, raw depth, RGB image, and our predicted depth image.}
	\label{intro}
\end{figure}

These characteristics are significant in the frequency domain and can be effectively utilized for depth completion tasks. For single-channel depth image $D\in{H\times W}$, the Discrete Fourier Transform is defined as

\begin{equation}
    S(D)(u,v) = \sum_{h=0}^{H-1} \sum_{w=0}^{W-1} D(h,w) e^{-j2\pi (\frac{h}{H}u+\frac{w}{W}v)}  
\end{equation}
where $u$ and $v$ denote the coordinates of the Fourier space. This process transforms a spatial depth map into complex frequency-domain coefficients, illustrating the spectral properties of images. The Fourier image can be divided into real part $\mathcal{R}(D)(u,v)$ and imaginary part $\mathcal{I}(D)(u,v)$, leading to the amplitude $A(D)(u,v)$ and phase $\varphi(D)(u,v)$ spectra: 

\begin{align}
    A(D)(u,v) &= \sqrt{{\mathcal{I}(D)(u,v)}^2 + {\mathcal{R}(D)(u,v)}^2} \\
    \varphi(D)(u,v) &= \arctan[\frac{\mathcal{I}(D)(u,v)}{\mathcal{R}(D)(u,v)}]
\end{align}


As shown in Fig.~\ref{intro}, the Fourier-transformed image decomposes visual information through two fundamental components: The amplitude spectrum quantifies the energy distribution across spatial frequencies, where lower frequencies correspond to gradual intensity variations and higher frequencies represent abrupt intensity changes. The phase spectrum preserves structural semantics by encoding the spatial relationships between frequency components.


The invalid depth regions exhibit a significantly lower density of data points compared to standard areas, resulting in an uneven sampling space. This degradation disrupts the spatial continuity of depth images and may affect both the Fourier amplitude spectrum and the phase spectrum.

In the Fourier domain amplitude spectrum, the sharp edges of invalid depth regions introduce additional high-frequency components, leading to an abnormal elevation of energy in high-frequency regions. Moreover, the presence of invalid regions disrupts the general smoothness of the scene structure, causing an abnormal decrease in low-frequency components.

Regarding the Fourier domain phase spectrum, uneven sampling space disrupts phase relationships among different frequency components, leading to phase inconsistencies resembling aliasing effects or stochastic noise. Since phase information predominantly encodes structural and spatial positioning details, the phase values within invalid depth areas fail to correctly represent spatial relationships. 

Based on these observations, this method is motivated by the physical characteristics behind the spectral degradation in the invalid areas. Our strategy utilizes low-frequency components to restore the overall shape of invalid regions, while high-frequency filtering mitigates ringing artifacts along invalid area boundaries. This ensures a balance between detail preservation and artifact suppression, leading to reconstructed depth structures that maintain spatial continuity with well-defined edges.

\subsection{General Architecture}

\begin{figure*}[t]
	\centering
	\includegraphics[width=\linewidth,scale=1.00]{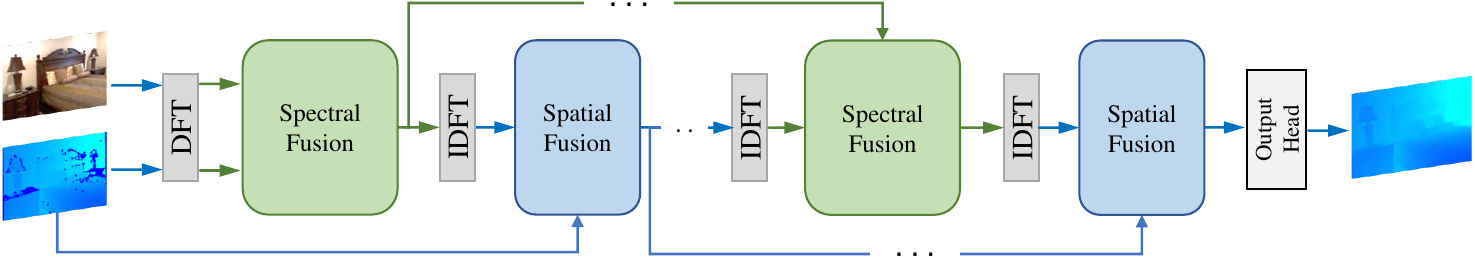}
	\caption{Overview of our S2ML method. Given $D^\mathrm{raw}$ and $I$ as input, they are initially embedded into feature representations. These features undergo a recursive fusion process through a series of spatio-spectral fusion pairs. DFT and IDFT are conducted to enable information interaction between the frequency domain and the spatial domain. Subsequent to the frequency fusion, the fused frequency features are conveyed directly to the ensuing frequency fusion module and concurrently transformed into the spatial domain to be processed by the spatial fusion module.}
	\label{model architecture}
\end{figure*}

Existing image-guided depth completion methods typically operate in the spatial domain, extracting features from both modalities and reconstructing the depth map accordingly. In contrast, we harness the potential complementary information available in the frequency domain and design our model architecture to perform multi-modal fusion in the frequency domain. As depicted in Fig.~\ref{model architecture}, our proposed S2ML comprises two major modules: the spatial fusion module and the spectral fusion module, which includes both phase fusion and amplitude fusion. Particularly, the frequency branch is capable of performing global analysis, as each coefficient in the frequency domain encapsulates information from the entire image.

For the input raw depth image $D^\mathrm{raw}$ and RGB image $I$, we initially embed them using ResNet backbone to obtain the initial depth feature map and RGB image feature map. Subsequently, these feature maps are fed into the spectral fusion module for frequency-domain feature fusion. The resulting spectral feature map is then combined with the original embedded depth feature and input into the spatial fusion module to refine the depth map. The outputs of both the spatial fusion module and the frequency fusion module are further fed into the next frequency fusion module.

Through the cascaded spatio-spectral fusion pairs, the depth map is progressively refined by leveraging the complementary information from the RGB image in both the frequency domain and the spatial domain. Finally, the output depth feature map from the last spatial fusion module is forwarded to the depth output head to predict the final depth completion result.

\subsection{Frequency Fusion Module}

\begin{figure}[t]
	\centering
	\includegraphics[width=\linewidth,scale=1.00]{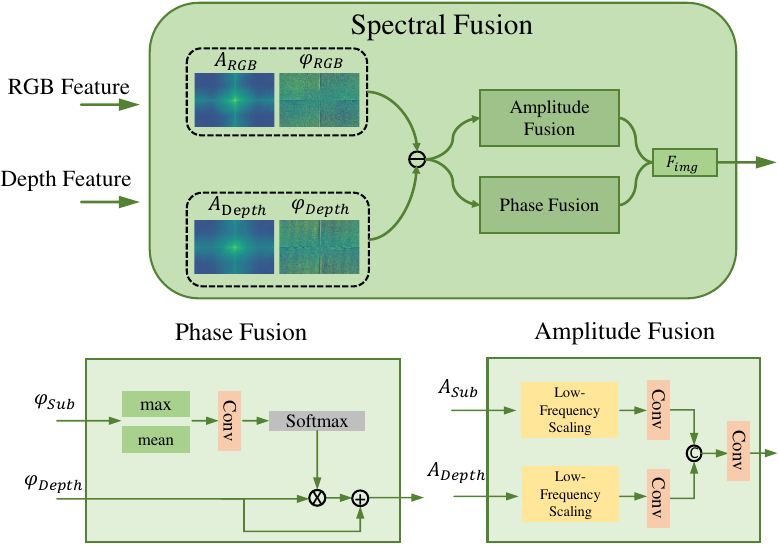}
	\caption{Structure of our proposed frequency fusion module, involving the fusion of amplitude and phase spectra from two modalities through distinct fusion strategies. To underscore the differential information present within the frequency domain, the module extracts spectrum difference feature maps between the depth spectrum and the RGB spectrum, facilitated by a residual connection from the depth spectrum. This approach guides the network to prioritize spectral discrepancies during the fusion process.}
	\label{frefuse}
\end{figure}

In the frequency domain, images consist of the amplitude and phase components, representing energy distribution and semantic information, respectively. Existing frequency-domain image processing methods often neglect the distinct roles of the amplitude and phase information. Typically, these methods focus on image restoration tasks such as deraining, deblurring, super-resolution, and dehazing, where the degradation process primarily affects the amplitude spectrum while leaving the phase unchanged. Consequently, their approaches often concentrate solely on amplitude utilization or treat both amplitude and phase equivalently, thereby overlooking the full potential of information available in the frequency domain.

Our approach introduces a novel spectral fusion method that distinctly processes the phase and amplitude information. As depicted in Fig.~\ref{frefuse}, the $i$-th RGB feature map $F_\mathrm{RGB}^i$ and depth feature map $F_\mathrm{depth}^i$ are initially transformed into the frequency spectra:

\begin{align}
    S_\mathrm{RGB}^i &= \mathrm{DFT}(F_\mathrm{RGB}^i) \\
    S_\mathrm{depth}^i &= \mathrm{DFT}(F_\mathrm{depth}^i)
\end{align}

These spectra are then decomposed into their phase and amplitude components:

\begin{align}
    \varphi_\mathrm{RGB}^i , A_\mathrm{RGB}^i &= f_d(S_\mathrm{RGB}^i)\\
    \varphi_\mathrm{depth}^i , A_\mathrm{depth}^i &= f_d(S_\mathrm{depth}^i)
\end{align}
where $f_d$ denotes the spectrum decomposition function, $\varphi_\mathrm{RGB}^i$ and $A_\mathrm{RGB}^i$ stand for the phase and amplitude of $F_\mathrm{RGB}^i$, respectively, and $\varphi_\mathrm{depth}^i$ and $A_\mathrm{depth}$ represent the phase and amplitude of $F_\mathrm{depth}^i$. 

To leverage the complementary information from both modalities, we extract two spectrum difference features from the phase and amplitude spectra, respectively, between the depth and RGB images: $\varphi_\mathrm{sub}^i = |\varphi_\mathrm{depth}^i - \varphi_\mathrm{RGB}^i|$ and $A_\mathrm{sub}^i = |A_\mathrm{depth}^i - A_\mathrm{RGB}^i|$. Based on our previous analysis of how various degradation patterns affect high- and low-frequency components, it is clear that accurately reconstructing the high-frequency part is essential for effective depth completion. Consequently, we adjust the low-frequency component of the amplitude image accordingly:

\begin{align}
    A_\mathrm{depth}^{*i} &= \alpha * M * A_\mathrm{depth}^i + (1-M) * A_\mathrm{depth}^i \label{eq:dep}\\ 
    A_\mathrm{sub}^{*i} &= \alpha * M * A_\mathrm{sub}^i + (1-M) * A_\mathrm{sub}^i \label{eq:sub}
\end{align}
where $M$ is the low-frequency mask and $\alpha$ is the weight of the low-frequency component. The amplitude images are then concatenated in the channel dimension to conduct band-wise fusion:

\begin{equation}
    A_f^i = f_c(\mathcal{C}(A_\mathrm{depth}^{*i},A_\mathrm{sub}^{*i})) + A_\mathrm{depth}^i
\end{equation}
where $A_f$ denotes the fused amplitude image and $f_c$ denotes the convolution function. The amplitude spectrum, representing the intensity information of each frequency component, is reconstructed by extracting band-wise amplitude information from both modalities based on the physical priors of degradation on low- and high-frequency parts.

The semantic details within the phase spectrum can serve as complementary information for invalid areas. Due to the absence of explicit correlations between physical priors and band-wise properties in the phase spectrum, 
we use the difference phase spectrum $\varphi_\mathrm{sub}$ to calculate a pixel-wise attention score in channel dimension. This score emphasizes the semantics of different frequency components and modulates the depth phase spectrum $\varphi_{depth}$:

\begin{align}
    \varphi_\mathrm{sub}^{*i} =  f_c(\mathcal{C}(mean({\varphi_\mathrm{sub}^i}),max(\varphi_\mathrm{sub}^i))
\end{align}
where the $mean$ and $max$ represent the channel-wise average and maximum of $\varphi_\mathrm{sub}$, respectively. The coefficient matrix is calculated by the Sigmoid function and used to modulate the depth phase image:

\begin{align}
    \varphi_f^i = Sigmoid(\varphi_\mathrm{sub}^{*i}) * \varphi_\mathrm{depth}^i + \varphi_\mathrm{depth}^i
\end{align}
where $\varphi_f$ denotes the fused phase image. With the fused amplitude image and phase image, we can reconstruct the fused spectrum:

\begin{equation}
    S_\mathrm{fused}^i = f_r(\varphi_f^i , A_f^i)
\end{equation}
here, $f_r$ denotes the spectrum reconstruct function. The fused spectrum is then fed into subsequent modules.

\subsection{Image Fusion Module}

\begin{figure}[t!]
	\centering
	\includegraphics[width=\linewidth,scale=1.00]{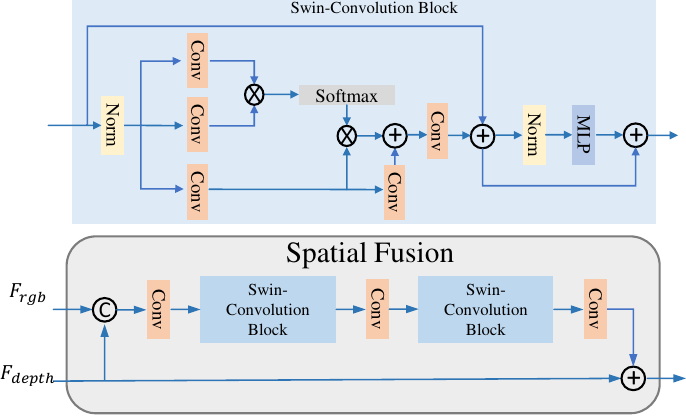}
	\caption{Structure of our proposed image fusion module. This module combines convolutional layers and a Swin-Transformer architecture to extract both global and local features from the input images. The Swin-Transformer excels at capturing long-range dependencies, while convolutional layers handle local details. The details of window partitioning and merging within the Swin-Transformer are omitted for brevity. }
	\label{imgfuse}
\end{figure}

As illustrated in Fig.~\ref{imgfuse}, the spatial fusion module processes the spectral features and the spatial features. The fused spectral features are transformed back to the spatial domain to align with the spatial features:

\begin{equation}
    F_{sf}^i = \mathrm{IDFT}(S_\mathrm{fused}^i)
\end{equation}
where the $F_{sf}^i$ denotes the spatial feature map of the fused spectrum obtained by performing IDFT. Then the feature maps from two modalities are concatenated and fed into a $1\times 1$ convolutional layer:

\begin{equation}
    F_\mathrm{in}^i = f_c(\mathcal{C}(F_{sf}^i,F_\mathrm{depth}^i))
\end{equation}
where $F_\mathrm{in}^i$ denotes the aggregated feature map. Inspired by~\cite{zhang_depth_2024}, We employ a combination of multiple Swin-Transformer blocks and convolutional layers to extract both local and global features from the feature map. The Swin-Transformer blocks enhance the quality of depth reconstruction by capturing contextual relationships, as different regions of the depth map may exhibit similar depth values. Meanwhile, convolutional layers effectively identify clear edges between objects, which is essential for precise depth completion. By integrating Swin-Transformer modules (for global context) with convolutional layers (for local details), we achieve a comprehensive refinement of the depth map that addresses both local and global aspects.

Specifically, as illustrated in Fig.~\ref{imgfuse}, we integrate self-attention with convolution by using convolutional layers to extract the query, key, and value matrices, which are then used to compute the attention map. This attention map is combined with the value matrix to merge both local and global information. The attention map is computed using Swin-Transformer techniques, including window partitioning and window merging. To progressively refine the depth feature map, we cascade multiple Swin-Convolution blocks. Therefore, the $j$-th feature map $F_{j}^i$ extracted by the Swin-Convolution module is defined as follows:

\begin{equation}
    F_{j}^i = f_{\mathrm{Swin}\text{-}C}^j (F_{j-1}^i),   j= 1,2,3,...,J
\end{equation}
where $f_{\mathrm{Swin}\text{-}C}^j$ denotes the $j$-th Swin-Convolution block, and $F_{0}^i = F_\mathrm{in}^i$. The final feature map $F_{J}^i$ from the Swin-Convolution blocks is combined with the depth feature map and then passes through a $3\times 3$ convolutional layer to obtain $F_\mathrm{out}^i$. 

\begin{equation}
    F_\mathrm{out}^i = f_c(F_{J}^i + F_\mathrm{depth}^i)
\end{equation}

The fused feature map of the last spatial fusion module is fed into the depth head to obtain the depth completion result.

\begin{figure}[h]
	\centering
	\includegraphics[width=\linewidth,scale=1.00]{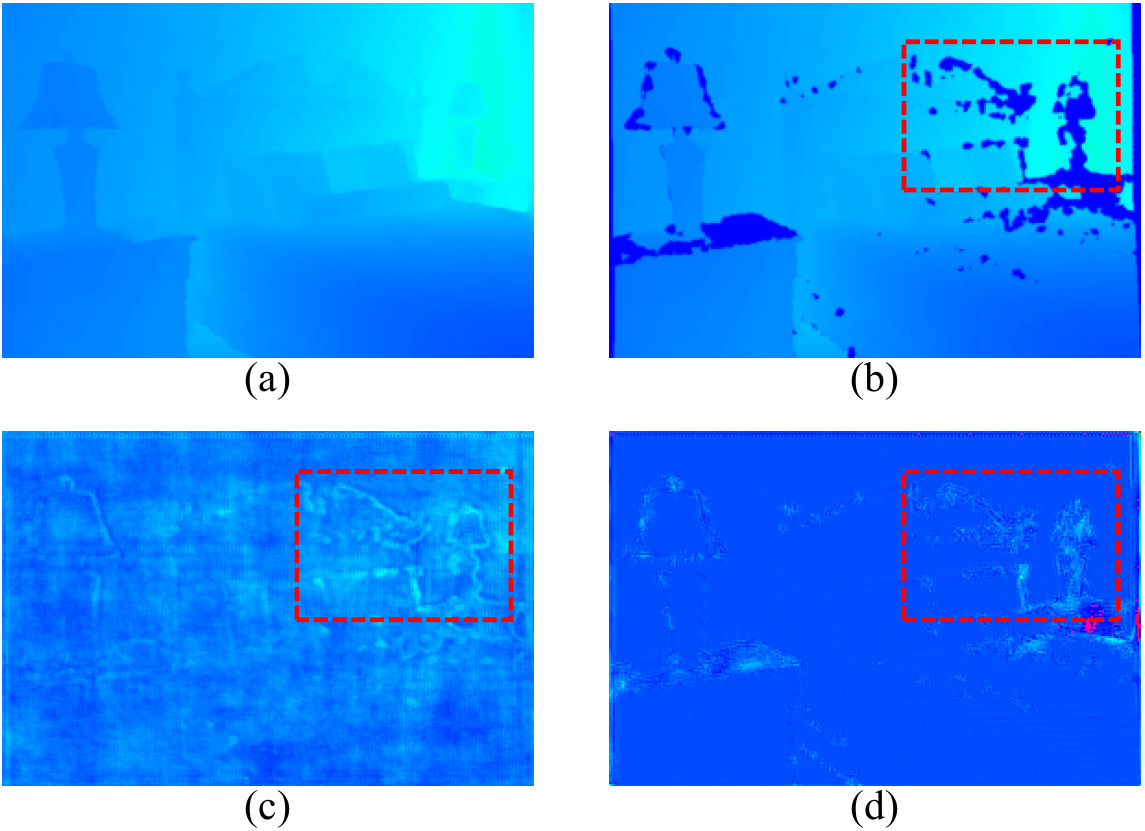}
	\caption{Visualization of depth feature maps: (a) Ground truth, (b) Raw depth map, (c) Residual depth feature map of a single frequency fusion module, (d) Residual depth feature map of a spatio-spectral fusion pair. The invalid areas and corresponding residuals within the red rectangles highlight the contribution of each module to the depth completion process. }
	\label{middle}
\end{figure}

The effectiveness of the spatio-spectral fusion module can be visualized through the residual depth feature maps. For the $i$-th spatio-spectral fusion pair, the residual feature map of spectral fusion module $R_{FF}^i$ and the residual feature map of spatio-spectral fusion pair $R_{FI}^i$ can be calculated by
\begin{align}
   R_{FF} = f_\mathrm{head}(F_{sf}^i - F_\mathrm{depth}^i) \\
   R_{FI} = f_\mathrm{head}(F_\mathrm{out}^i - F_\mathrm{depth}^i)
\end{align}
where $f_\mathrm{head}(\cdot)$ represents the output head of our depth completion model, which is used to visualize the high-dimensional feature maps. As illustrated in Fig.~\ref{middle}, we present the feature maps $R_{FF}$ in Fig.~\ref{middle}(c) and $R_{FI}$ in Fig.~\ref{middle}(d). From Fig.~\ref{middle}(c), it is evident that our proposed spectral fusion module focuses on reconstructing features in the invalid areas and emphasizes restoring clear edges. In (d), we demonstrate the overall depth completion capability of the spatio-spectral fusion pair. This approach enhances the quality and reliability of the reconstructed depth image.

\subsection{Loss Function}
We apply both $\mathcal{L}_1$ and $\mathcal{L}_2$ loss functions  to improve the depth completion performance, as defined below:

\begin{align}
    \mathcal{L}_\mathrm{all}(D^\mathrm{pred},D^\mathrm{GT}) = \gamma_1 \mathcal{L}_1(D^\mathrm{pred},D^\mathrm{GT})  + \gamma_2 \mathcal{L}_2(D^\mathrm{pred},D^\mathrm{GT}) 
\end{align}
where
\begin{align}
    &\mathcal{L}_n = \frac{1}{P} \sum_i^W \sum_j^H |D^\mathrm{pred}_{i,j} - D^\mathrm{GT}_{i,j} \cdot \mathbb{I} ({D^\mathrm{GT}_{i,j}>0}) |^{n} \\
    &P = \sum_i^W \sum_j^H \mathbb{I} ({D^\mathrm{GT}_{i,j}>0})
\end{align}
where $D^\mathrm{GT}$ is the ground truth depth image, $D^\mathrm{GT}_{i,j}$ and $D^\mathrm{pred}_{i,j}$ represent the pixel in the ground truth depth image and the predicted depth image. $\mathbb{I} ({D^\mathrm{GT}_{i,j}>0})$ is the indicator function for the validity of the pixel in the ground truth depth image. $\gamma_1$,$\gamma_2$ =1.0, $n \in \{1,2\}$.

\section{Experimental Results}
\subsection{Datasets and Metrics}
We conducted experiments on two widely-used benchmarks: NYU-Depth v2 ~\cite{silberman_indoor_2012} and SUN RGB-D ~\cite{song_sun_2015}.

\textbf{NYU-Depth v2.} The NYU-Depth v2 dataset is a predominant indoor depth completion dataset. It comprises RGB and depth image pairs collected using Kinect across 464 scenes. Following the established protocol, we split the densely labeled image pairs into a testing set of 654 pairs and a training set of 795 pairs. Each pair includes a raw depth image from sensors, a reconstructed depth image, an RGB image, and a segmentation mask. The input images were first resized to $240 \times 320$ and then center cropped to $192 \times 288$. 

\textbf{SUN RGB-D.} The SUN RGB-D dataset contains 10,335 RGB-D images captured by four different sensors. This dataset includes samples from various scenes and devices, which is beneficial for assessing the model's generalization ability. We used 4,807 image pairs for training and 4,313 image pairs for testing. The input images were resized to $240 \times 320$ and then center cropped to $192 \times 288$. 

\subsection{Metrics}
Following the NYU-Depth v2 benchmark and existing depth completion methodologies, we utilized three standard metrics for evaluating depth completion performance: root mean square error (RMSE), mean absolute relative error (REL), and $\delta_{i}$, which represents the percentage of predicted depth pixels falling within the thresholds $i$. 

\subsection{Comparison with SOTAs}

To evaluate the performance of our proposed S2ML method, we compared it against several state-of-the-art (SOTA) deep learning-based methods. Our model was implemented with optimal parameter settings derived from our ablation studies. Performance measurements of competing methods were obtained through experiments on the NYU-Depth v2 and SUN RGB-D datasets using their default model settings. All methods were trained for the raw depth map completion task and tested under the same protocol.

\begin{figure*}[h]
	\centering
	\includegraphics[width=\linewidth,scale=1.00]{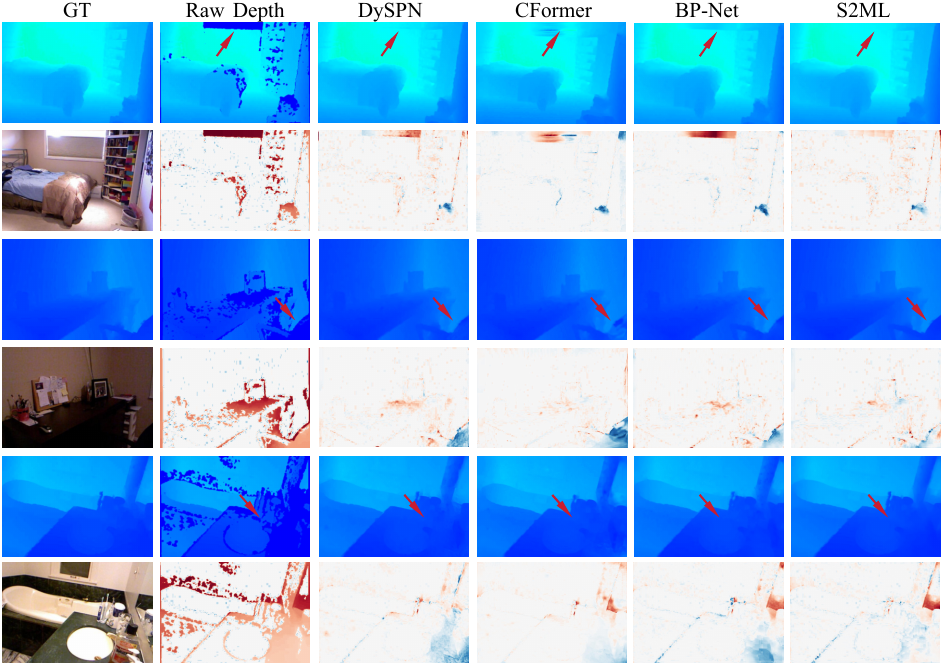}
	\caption{Qualitative depth completion comparison results and their corresponding residual maps of different methods on NYU-Depth v2. The positive values are represented by varying shades of blue, while the negative values are represented by varying shades of red.}
	\label{compare}
\end{figure*}

\textbf{NYU-Depth v2.} We first evaluate our method on the widely used NYU-Depth v2 dataset. Typical images from this dataset and the corresponding completion results are depicted in Fig.~\ref{compare}, with quantitative results shown in Table~\ref{tab:1}. 

In the qualitative comparison, we present both the completion results and their corresponding residual images, colorized using red and blue shades to highlight the differences. We focus on invalid depth regions and their corresponding edges that pose significant reconstruction challenges. It is evident that our proposed S2ML produces the most distinct and clear edges in the predicted depth images, indicating its high efficacy in boundary delineation, which is crucial for accurate depth perception. Additionally, the residual images with our method contain the least residues compared to the other methods, suggesting that our method is effective in minimizing reconstruction errors and preserving the semantic integrity of the original scene. This superior performance is attributed to the integration of spectral images, featured with rich edge and semantic information, and is further enhanced through spatial fusion at both the local and global levels.

In the quantitative comparison, our method outperforms all the other baselines by a large margin in RMSE and exceeds them by at least $9.1\%$ in REL. Our method also achieves the best score in $\delta_{i}$, reaching 100 \% at the threshold $i = 1.25^3$. Given that NYU-Depth v2 is the most widely used benchmark for indoor depth completion, our experimental results on this dataset demonstrate that S2ML surpasses the compared SOTA methods in overall performance. The superiority of our method can be primarily attributed to two key innovations: the spectral fusion strategy and the global-local spatial fusion paradigm. These components enhance the depth completion process by effectively integrating both spatial and spectral information, resulting in improved accuracy and robustness compared to existing methods.

\begin{table*}[h]
        \centering
	\caption{Quantitative comparison with other methods on the NYU-Depth V2}
	\label{tab:1}  
        \setlength{\tabcolsep}{12pt} 
	\begin{tabular}{c|c|c|ccc|c|c}
		\hline
		Method & RMSE$\downarrow$ & REL$\downarrow$ & $\delta_{1.25}$$\uparrow$& $\delta_{1.25^2}$$\uparrow$& $\delta_{1.25^3}$$\downarrow$ & FLOPs(G) $\downarrow$& Parameters(M) $\downarrow$ \\
		\hline\hline
		RDF-GAN ~\cite{wang_rgb-depth_2022}& 0.139 & 0.013 & 98.7 & 99.6 & 99.9 & 177.26 & 93.0  \\
		AGG ~\cite{chen_agg-net_nodate}    & 0.092 & 0.014 & 99.4 & 99.9 & 100.0 & 95.6 & 92.4 \\
            DySPN ~\cite{lin_dynamic_2022}  & 0.091 & 0.011 & 99.4 & 99.0 & 100.0 & \textbf{55.9} & \textbf{16.5} \\
            GraphCSPN ~\cite{avidan_graphcspn_2022} & 0.099 & 0.015 & 99.0 & 99.8 & 99.9 & 93.0 & 89.9 \\
            CFormer ~\cite{zhang_completionformer_2023} & 0.088 & 0.011 & 99.4 & 99.9 & 100.0 & 389.4 & 45.8 \\
            BP-Net ~\cite{tang_bilateral_2024} & 0.119 & 0.015 & 99.0 & 99.8 & 99.9 & 137.1 & 89.9 \\
            GuideFormer~\cite{rho_guideformer_2022} & 0.126 & 0.016 & 99.1 & 99.8 & 99.9 & 73.3 & 25.1  \\
            LRRU~\cite{wang2023lrru} & 0.091 & 0.011 & 99.6 & 99.9 & 100.0 & 77.4 & 21.0  \\ 
            S2ML & \textbf{0.080} & \textbf{0.010} & \textbf{99.5} & \textbf{99.9} & \textbf{100.0} & 207.8 & 28.4 \\
	       \hline
	\end{tabular}
\end{table*}

\textbf{SUN RGB-D.} We also conducted experiments on the large-scale SUN RGB-D dataset, which comprises samples collected using various hardware devices in diverse scenes. These experiments are crucial for evaluating the generalizability of our method. {As shown in Table~\ref{tab:2}, our proposed S2ML outperforms the other SOTA methods in terms of most metrics. Since the SUN RGB-D dataset is composed of data collected from diverse devices, the experiments conducted on this dataset demonstrate the generalizability of our approach across different hardware configurations. (The results of BP-Net are not included due to its significant dependence on camera intrinsic parameters, which results in its inability to converge in scenarios involving cross-hardware device conditions.)}

As quantitatively demonstrated in Table~\ref{tab:1}, our method achieves the best depth completion accuracy across all metrics while maintaining competitive computational efficiency. \textcolor{red}{The incorporation of spectral fusion modules results in a marginal increase in computational cost, introducing an additional 0.1G FLOPs and 1.0M parameters. Consequently, the total computational complexity of the proposed model amounts to 28.4M parameters and 207.8G FLOPS. Despite this increment, the overall parameter count of the proposed model remains modest, approximately 30\% of that of the state-of-the-art BP-Net, which contains 89.9M parameters.} Compared to the lightest spatial method DySPN, our approach delivers $12.1\%$ better RMSE and $9.1\%$ better REL with $11.9M$ parameter increase. While outperforming attention-based CompletionFormer by $9.1\%$ in RMSE and $9.1\%$ in REL, our method requires $46.6\%$ fewer parameters and $46.7\%$ less computation. This efficiency stems from our Fourier domain module design, which effectively extracts complementary information in the frequency domain without relying on cascaded neural networks. Additionally, the iterative interaction between spatial and spectral information enhances reconstruction efficiency, as the Swin-Convolution module is adept at integrating the complementary features extracted from the frequency domain while maintaining a relatively small model size.

\begin{table}[t!]
        \centering
	\caption{Quantitative comparison with other methods on SUN RGB-D}
	\label{tab:2}  
	\begin{tabular}{c|c|c|ccc}
		\hline
		Method & RMSE$\downarrow$ & REL$\downarrow$ & $\delta_{1.25}$$\uparrow$& $\delta_{1.25^2}$$\uparrow$& $\delta_{1.25^3}$$\downarrow$   \\
		\hline\hline
		RDF-GAN & 0.263 & 0.058 & 96.3 & 98.3 & 99.0  \\
		AGG     & 0.154 & 0.039 & 98.4 & 99.0 & 99.4\\
            DySPN   & 0.195 & 0.080 & 97.5 & 98.9 & 99.4  \\
            GraphCSPN & 0.135 & 0.042 & 98.8 & 99.4 & 99.7\\
            CFormer & 0.131 & 0.032 & 98.4 & 99.4 & 99.6 \\
            GuideFormer & 0.147 & 0.043 & 98.2 & 99.3 & 99.4 \\
            LRRU & 0.121 & 0.034 & 99.0 & 99.3 & 99.5 \\
            S2ML & \textbf{0.119} & \textbf{0.030} & \textbf{99.0} & \textbf{99.4} & \textbf{99.6} \\
	       \hline
	\end{tabular}
\end{table}

Overall, the experimental results demonstrate that our S2ML establishes a new strong baseline for depth completion. The two major components of our network, the spectral fusion module, and the spatial fusion module, significantly enhance the overall depth completion performance, enabling our method to surpass existing techniques across popular benchmarks and metrics. Our method demonstrates inherent robustness to common RGB degradations through principled spectral processing. Since image noise and poor lighting primarily distort the amplitude spectrum (particularly high-frequency components for noise, and low-frequency bands for illumination variations), our frequency-aware fusion strategy dynamically adjusts the importance of different spectral components. By emphasizing phase coherence that preserves semantic structures while adaptively suppressing noise-contaminated high frequencies and recalibrating illumination-affected low frequencies, we maintain stable depth estimation under degradation. Crucially, the spectral fusion module complements this process by leveraging semantically intact features from the RGB domain, which remain resilient to such distortions as they primarily affect pixel-level intensities rather than high-level structural patterns. This dual-domain synergy enables our model to utilize RGB guidance effectively even with imperfect inputs, as evidenced by our superior performance across benchmarks under varying imaging conditions.

\subsection{Sensitivity Analysis}

{
In real-world scenarios, RGB images often suffer from issues such as occlusions, poor lighting, and noise. To assess the robustness of our method under such conditions, we conducted experiments using simulated RGB image degradations. Specifically, we simulated scenarios including occlusion, noise, extreme noise (noise+), poor lighting (PL), and extremely poor lighting (PL+). Our model was trained on datasets free from degradation and evaluated on datasets with degraded RGB images. Both qualitative and quantitative analyses were performed to thoroughly evaluate the robustness and effectiveness of our framework.}

\begin{figure*}[t]
	\centering
	\includegraphics[width=\linewidth,scale=1.00]{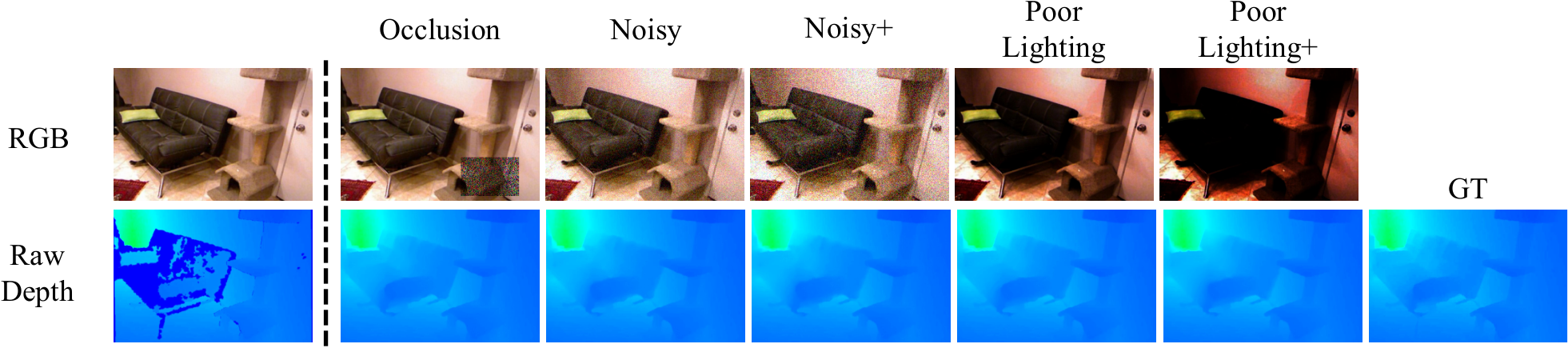}
	\caption{{Evaluation of robustness under various RGB image degradation. The reconstructed depth images exhibit remarkable consistency and reliability, even under conditions of extreme noise or poor lighting. These results underscore the effectiveness of our framework in handling challenging real-world scenarios.}}
	\label{sensitivity}
\end{figure*}

{
As shown in Fig.~\ref{sensitivity}, we tested our framework on RGB images subjected to five distinct degradation scenarios, all of which are commonly encountered in real-world applications. These tests are designed to assess the robustness and practical utility of our framework in realistic conditions. The second row of the figure displays the raw depth, reconstructed depth, and ground truth (GT) depth. It is evident that our network consistently produces clear and accurate depth predictions, even under varying degradation patterns, demonstrating its reliability in challenging environments.}

\subsection{Evaluation in Outdoor Environments}

\begin{figure*}[t]
	\centering
	\includegraphics[width=\linewidth,scale=1.00]{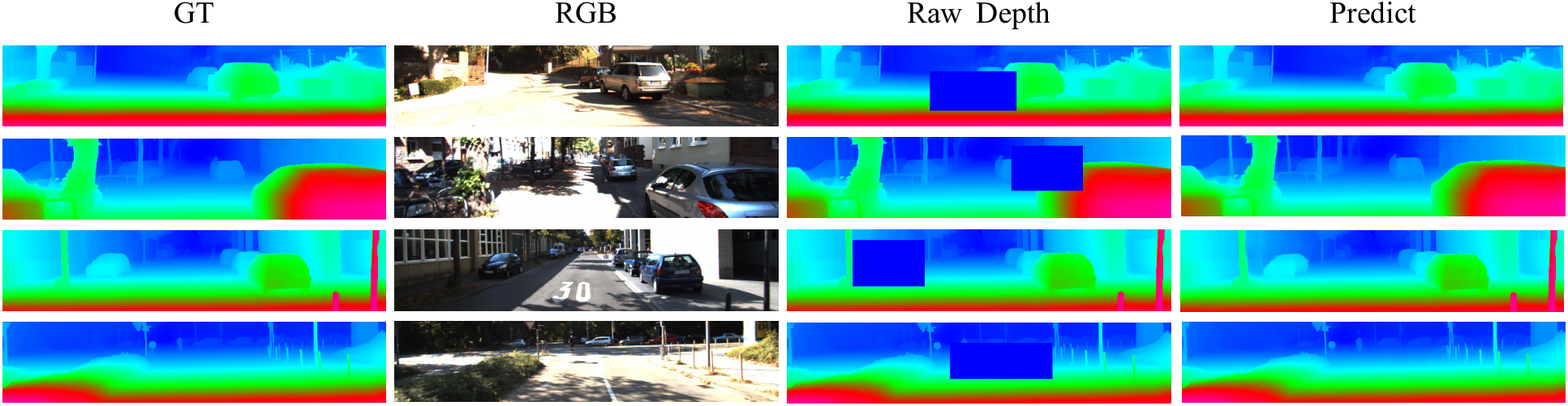}
	\caption{{Evaluation of robustness in outdoor environments. The GT depth images are estimated using the DepthAnything model, while the raw depth images were generated by randomly collapsing depth values to zero within a randomly selected area. This process was designed to simulate potential distortions or occlusions that might occur in real-world outdoor settings, thereby providing a rigorous test of the framework's robustness to such challenges.}}
	\label{kitti_img}
\end{figure*}

{To assess the robustness of our method in real-world scenarios across diverse environments, we performed depth completion evaluations on outdoor scenes. Due to the lack of dense outdoor depth image datasets, we generated simulated depth images using 1000 outdoor RGB images from the KITTI datasets~\cite{geiger2013vision} and synthesized their corresponding depth image using the DepthAnything~\cite{yang2025depth} model. To simulate potential sensor occlusions, random occlusions (15\% areas) are applied to these depth images, generating raw depth data and creating a controlled testbed for evaluating the framework’s robustness to such disturbances. The average RMSE between the raw depth and GT depth is 0.134, while the RMSE between the predicted depth and GT depth is 0.073 ( 45.5\% reduction), demonstrating the effectiveness of our method in outdoor scenarios. As illustrated in Fig.~\ref{kitti_img}, our depth completion method can accurately reconstruct depth images even in the presence of large occlusion areas.}

\subsection{Ablation Study}
We conduct ablation studies on the NYU-Depth v2 dataset to assess the effectiveness of the individual components of our method.

\subsubsection{Spectral Fusion}

Given that multi-modal spectral fusion is one of the major contributions of this work, we conduct comprehensive experiments on the spectral fusion module to evaluate its effectiveness. Specifically, we evaluate different versions of our method:

\begin{itemize}
    \item \textbf{Version-1}: The spectral fusion module is replaced with a multi-layer perception (MLP) operating in the spatial domain.
    \item  \textbf{Version-2}: Multi-modal features are first transformed into the frequency domain and then fused using an MLP.
    \item  \textbf{Version-3}: The amplitude fusion part is removed from our frequency fusion module.
    \item  \textbf{Version-4}: The phase fusion part is removed from our frequency fusion module.
    \item  \textbf{Version-5}: Amplitude and phase fusion are conducted using convolution, following the approach in previous work~\cite{wang_sgnet_2023}.
\end{itemize}

All other settings remain consistent with those described in the previous section.

\begin{table}[t!]

        \centering
	\caption{Quantitative comparison between different spectral fusion methods}
	\label{tab:4}  
        \resizebox{\linewidth}{!}{
	\begin{tabular}{c|c|c|ccc|c|c}
 
		\hline
  
		Version & RMSE$\downarrow$ & REL$\downarrow$ & $\delta_{1}$$\uparrow$& $\delta_{2}$$\uparrow$& $\delta_{3}$$\downarrow$  & FLOPs(G) & Params(M) \\
  
		\hline\hline
  
		V-1 & 0.092 & 0.012 & 99.2 & 99.7 & 99.9 & 211.9 & 28.5  \\
		V-2 & 0.100 & 0.013 & 99.1 & 99.7 & 99.9 & 211.9 & 28.5 \\
            V-3 & 0.096 & 0.012 & 99.2 & 99.8 & 99.9 & 206.9 & 28.4 \\
            V-4 & 0.103 & 0.013 & 99.1 & 99.7 & 99.9 & 207.8 & 28.4 \\
            V-5 & 0.089 & 0.013 & 99.3 & 99.7 & 99.8 & 208.7 & 28.4 \\
            S2ML & \textbf{0.080} & \textbf{0.010} & \textbf{99.5} & \textbf{99.9} & \textbf{100.0} & 207.8 & 28.4 \\
            
	       \hline
        
	\end{tabular}}
\end{table}

As shown in Tab.~\ref{tab:4}, the integration of frequency-domain fusion significantly enhances depth completion performance. Version-5, which employs spectral fusion as proposed in~\cite{wang_sgnet_2023}, outperforms Version-1, demonstrating the effectiveness of the spectral fusion strategy. Our complete method, however, significantly outperforms both Version-1 and Version-5. This improvement is attributed to our strategy, which leverages the distinct properties of phase and amplitude information, fully utilizing the complementary information from both modalities. Notably, only the complete method and Version-5 surpass Version-1, indicating that successful frequency-domain fusion requires the integration of both phase and amplitude information, and omitting either can result in a loss of crucial information and degrade performance.

To further understand the impact of different frequency-domain fusion strategies, we visualized the residual feature maps $R_{FF}$ and $R_{FI}$ for different versions. The $R_{FF}$ is the residual feature map of the spectral fusion module and the $R_{FI}$ is the residual feature map of the spatio-spectral fusion pair. Since Version-1 does not perform fusion in the frequency domain, it is omitted in this part. Additionally, the residual images are not normalized to standard value intervals to better highlight the changes.

\begin{figure*}[t]
	\centering
	\includegraphics[width=\linewidth,scale=1.00]{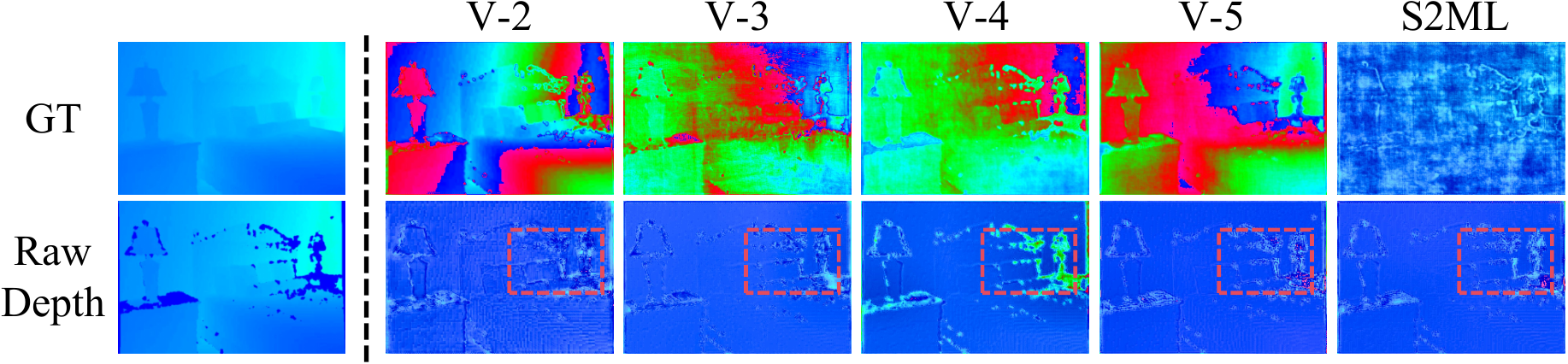}
	\caption{Residual feature maps $R_{FF}$ and $R_{FI}$ for different versions of our method. The $R_{FF}$ indicates the contribution and refinement of the spectral fusion module on the depth feature map, while the $R_{FI}$ illustrates the effect of the spatio-spectral fusion pair. V-2: The model replaces frequency fusion with an MLP. V-3: The model removes amplitude fusion. V-4: The model removes phase fusion. V-5: The model conducts frequency fusion using convolution.  }
	\label{abl}
\end{figure*}

As shown in Fig.\ref{abl}, the first row displays the feature residual maps $R_{FF}$, and the second row shows the $R_{FI}$. From the $R_{FF}$ maps, we observe that only the complete version of our method maintains stable value changes in the residual images. This stability is achieved by incorporating the RGB modality to modulate the depth frequency information during the fusion process, supplemented by a residual connection to further enhance stability. This property is crucial in spectral information fusion, as the spatial domain is highly sensitive to changes in the frequency domain. The $R_{FF}$ of Version-2 indicates that while spectral fusion provides some semantic and edge information, the depth values are chaotic and incorrect. The $R_{FF}$ of Version-3 lacks correct edge information due to the absence of amplitude information, while the $R_{FF}$ of Version-4 contains incorrect semantics because of missing phase information. Version-5 emphasizes edges and maintains correct semantics in the invalid areas of raw depth but also introduces rapid changes in spectral information, which hinders its depth completion ability.

On the other hand, the $R_{FI}$ maps show the overall contribution of the spatio-spectral fusion pair to depth completion. Since these residual images are from the last spatio-spectral fusion pair, the changes are relatively moderate. As highlighted in the red rectangles, the $R_{FI}$ of Version-2 contains additional values outside the invalid area, which can lead to incorrect depth completion. The $R_{FI}$ of Version-3 shows insufficient depth completion due to the omission of crucial amplitude information. In Version-4, the $R_{FI}$ reveals a pattern similar to mismatched amplitude and phase spectra, likely caused by the loss of phase information. Version-5 achieves proper depth completion by considering both amplitude and phase information, though it still introduces some excessive values due to instability in frequency domain fusion.

Our complete method overcomes these limitations through a dedicated fusion strategy designed for both frequency and spatial domains. In terms of spectral information, our distinct fusion strategy effectively and stably combines the amplitude and phase spectra, outperforming both Version-5 and Version-2. Additionally, the spatial fusion module seamlessly integrates auxiliary information from spectral fusion, resulting in a refined depth map.

\begin{table}[t!]

        \centering
	\caption{Quantitative comparison between different spectral fusion methods}
	\label{tab:sens}  
        \resizebox{\linewidth}{!}{
	\begin{tabular}{c|c|c|ccc}
 
		\hline
  
		Version & RMSE$\downarrow$ & REL$\downarrow$ & $\delta_{1}$$\uparrow$& $\delta_{2}$$\uparrow$& $\delta_{3}$$\downarrow$ \\
  
		\hline\hline
            Clean & {0.080} & {0.010} & {99.5} & {99.9} & \textbf{100.0} \\
		  Occlusion & 0.0841 & 0.011 & 99.4 & 99.9 & 100.0   \\
		Noisy & 0.0846& 0.011 & 99.4 & 99.8 & 100.0  \\
            Noisy+ & 0.0909 & 0.011 & 99.4 & 99.8 & 100.0  \\
            PL& 0.896 & 0.011 & 99.4 & 99.8 & 99.9  \\
            PL+ & 0.0977 & 0.012 & 99.2 & 99.8 & 99.9 \\
              
	       \hline
	\end{tabular}}
\end{table}

As illustrated in Table.~\ref{tab:sens}, compared with the test result on clean RGB images, the results on degraded images reveal the strong robustness of our method under various situations. This can be attributed to the design of phase fusion in the spectral fusion module that utilizes the complementary semantic information from RGB modality, which is stable when encountering these degradations.

\subsubsection{Different Number of spatio-spectral fusion pairs $\textit{N}$}

In our fusion model, the fused depth feature map undergoes recursive refinement through $N$ pairs of spatio-spectral fusion modules, as illustrated in Table.~\ref{tab:5}. To assess the impact of varying values of $N$, we conducted experiments on the NYU-Depth v2 dataset and evaluated the corresponding performance. All other experiment settings and metrics remained consistent with previous evaluations.

\begin{table}[t]

        \centering
	\caption{Quantitative comparison between different spatio-spectral fusion pairs number $\textit{N}$ settings}
	\label{tab:5}  
        
	\begin{tabular}{c|c|c|ccc}
 
		\hline
  
		$\textit{N}$ & RMSE$\downarrow$ & REL$\downarrow$ & $\delta_{1}$$\uparrow$& $\delta_{2}$$\uparrow$& $\delta_{3}$$\downarrow$  \\
  
		\hline\hline
  
		$\textit{N}=1$ & 0.098 & 0.013 & 99.3 & 99.8 & 99.9 \\
		$\textit{N}=2$ & \textbf{0.080} & \textbf{0.010} & \textbf{99.5} & \textbf{99.9} & \textbf{100.0} \\
            $\textit{N}=3$ & 0.084 & 0.011 & 99.4 & 99.8 & 100.0 \\
            $\textit{N}=4$ & 0.090 & 0.013 & 99.3 & 99.8 & 99.9 \\

	       \hline
        
	\end{tabular}
\end{table}

\begin{figure*}[t]
	\centering
	\includegraphics[width=\linewidth,scale=1.00]{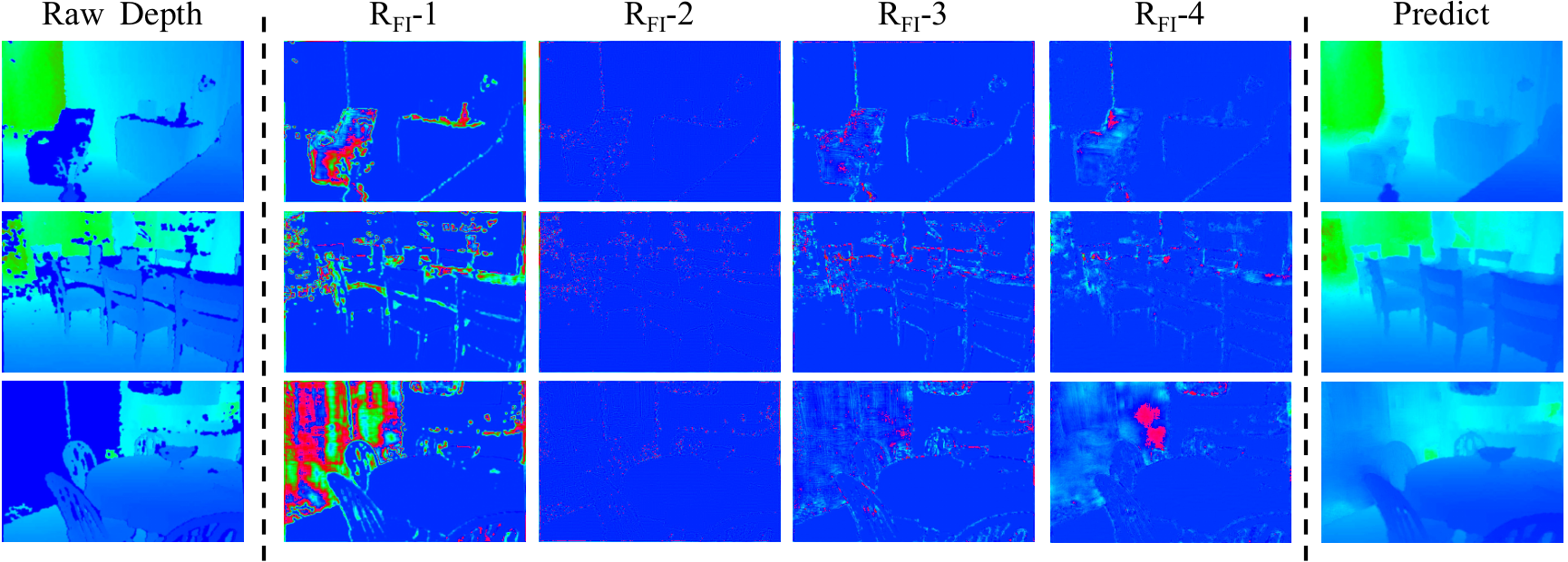}
	\caption{Residual feature maps $R_{FI}$ of each spatio-spectral fusion pair in model $N=4$(from the first to the last), which reveal the contribution of each pair in the depth completion process.   }
	\label{num}
\end{figure*}

As shown in Tab.~\ref{tab:5}, the choice of $N$ significantly affects the overall performance of our depth completion model. As anticipated, $N=1$ fails to provide adequate refinement for the depth feature map, resulting in suboptimal performance. This is evident from the higher error metrics and lower accuracy compared to other $N$ values. Notably, $N=2$ achieves the best performance among all tested values, thus being adopted as the configuration for our proposed model. However, increasing the depth to three and four layers leads to a slight performance decline, likely due to the added complexity, which can pose optimization challenges and increase the risk of overfitting.

To further understand the refining process between each spatio-spectral fusion pair and the performance degradation in deeper networks, we analyzed $R_{FI}$ for each fusion pair in the model with $N=4$. As illustrated in Fig.~\ref{num}, the first fusion pair exhibits a rapid change in the invalid area, with subsequent pairs gradually refining it to the correct depth value. However, in the model with $N=4$, it is evident that the contribution of the second fusion pair to the refinement of the depth map is minimal. This phenomenon is likely attributable to overfitting resulting from excessive network depth. Therefore, selecting an appropriate network depth is critical to achieving optimal performance. Based on our comprehensive analysis, we selected a configuration with $N=2$ to balance performance and complexity for our specific scenario.

{To determine the optimal number of spatio-spectral fusion pairs, Bayesian optimization~\cite{onorato2024bayesian} can be employed for network hyperparameter selection, enabling the identification of the ideal network depth. This optimization process can be seamlessly integrated into the existing workflow. Furthermore, experimental results indicate that our method demonstrates relative stability, as models with $N=2$ and $N=3$ both achieve strong performance.}

{In practical applications, datasets containing more diverse scenes and exhibiting more severe depth degradation may require deeper networks. To assess this, cross-validation can be incorporated by partitioning the dataset into multiple folds and evaluating network performance across different splits. This strategy allows for a more stable performance assessment, reducing bias associated with a single validation set. The optimal depth is then determined by aggregating performance metrics across folds, achieving a well-balanced trade-off between model expressiveness and generalization capability.}

{By leveraging these strategies, we can determine the appropriate number of network modules based on factors such as resolution ratio, scene complexity, and the proportion of invalid regions in the depth image. More intricate scenes generally require deeper network architectures to capture fine-grained details effectively. }

\section{Conclusion}

We introduced a novel Spatio-Spectral Mutual Learning framework designed to enhance indoor scene depth completion. Our approach seamlessly integrates modality fusion and depth feature map refinement across both frequency and spatial domains. Central to our method is a specialized spectral fusion module that leverages insights into the physical properties of invalid areas in raw depth images and their frequency domain characteristics. To fully harness the complementary spectral information from both modalities, we employed a heterogeneous fusion paradigm for amplitude and phase spectra. For amplitude fusion, we rescaled low-frequency components according to their significance, ensuring that crucial information is preserved. In phase fusion, we used an attention-based mechanism to effectively modulate depth semantics. In the spatial domain, our framework combines Swin-Transformer blocks with convolutional layers. The Swin-Transformer captures global context, while convolutional layers focus on extracting local edge details, resulting in a comprehensive and robust depth completion process.
Our extensive experiments on the NYU-Depth v2 and SUN RGB-D datasets, across a variety of hardware devices, consistently demonstrate the superior performance of our method compared to leading depth completion approaches. Additionally, we have conducted ablation studies to validate the efficacy of our spectral fusion design and to evaluate the contributions of individual modules. These thorough evaluations confirm the robustness and effectiveness of our method in depth completion tasks.

\bibliographystyle{IEEEtran}
\bibliography{IEEEabrv,reference3}


\begin{IEEEbiography}[{\includegraphics[width=1in,height=1.25in,clip,keepaspectratio]{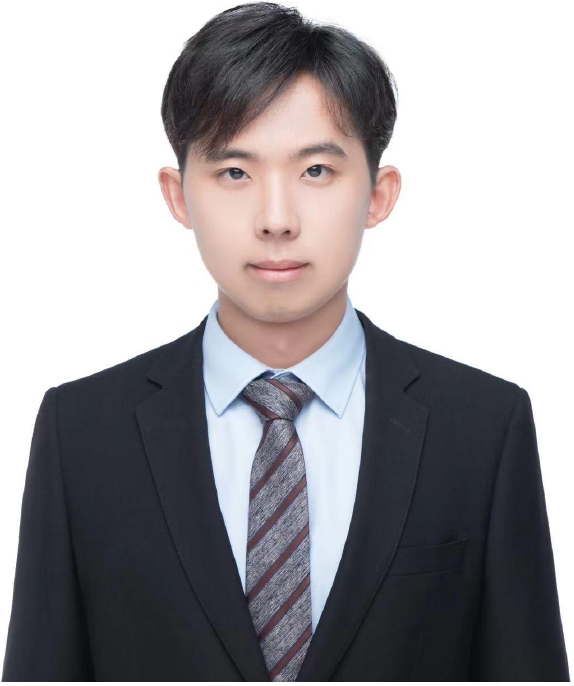}}] {Zihui Zhao} received a bachelor's degree from Beihang University in 2023. He is currently pursuing a Master's degree with the Institute of Data and Information, Tsinghua Shenzhen International Graduate School, Tsinghua University, China. His research interests include artificial intelligence and computer vision.
\end{IEEEbiography}


\begin{IEEEbiography}[{\includegraphics[width=1in,height=1.25in,clip,keepaspectratio]{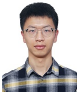}}] {Yifei Zhang} is currently a master's student at Tsinghua University. He obtained his bachelor's degree from Beihang University in 2022. His research interests include computational imaging and low-level vision.
\end{IEEEbiography}



\begin{IEEEbiography}[{\includegraphics[width=1in,height=1.25in,clip,keepaspectratio]{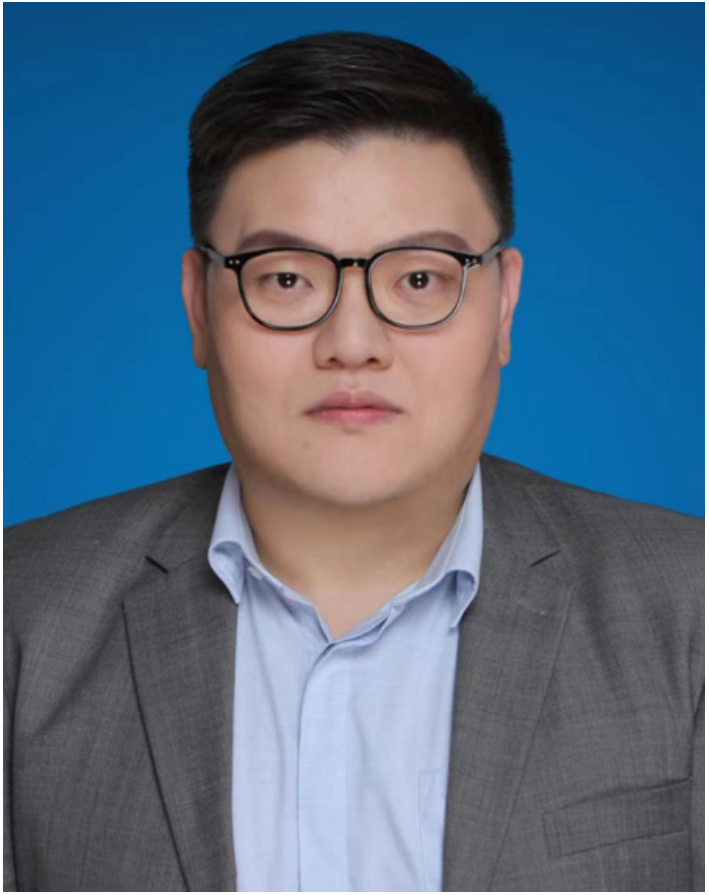}}] {Zheng Wang}(Senior Member, IEEE) received the B.S. and M.S. degrees from Wuhan University, Wuhan, China, in 2006 and 2008, respectively, and the Ph.D. degree from the National Engineering Research Center for Multimedia Software, School of Computer Science, Wuhan University, in 2017. He was a JSPS Fellowship Researcher with the Shin’ichi Satoh’s Laboratory, National Institute of Informatics, Japan, and a Project Assistant Professor with The University of Tokyo, Japan. He is currently a Professor with the National Engineering Research Center for Multimedia Software, School of Computer Science, Wuhan University. His research interests include multimedia content analysis and retrieval. He received the Best Paper Award from the 15th Paciffc-Rim Conference on Multimedia (PCM 2014), the 2017 ACM Wuhan Doctoral Dissertation Award, and the 2023 ACM Wuhan Rising Star.
\end{IEEEbiography}

\begin{IEEEbiography}[{\includegraphics[width=1in,height=1.25in,clip,keepaspectratio]{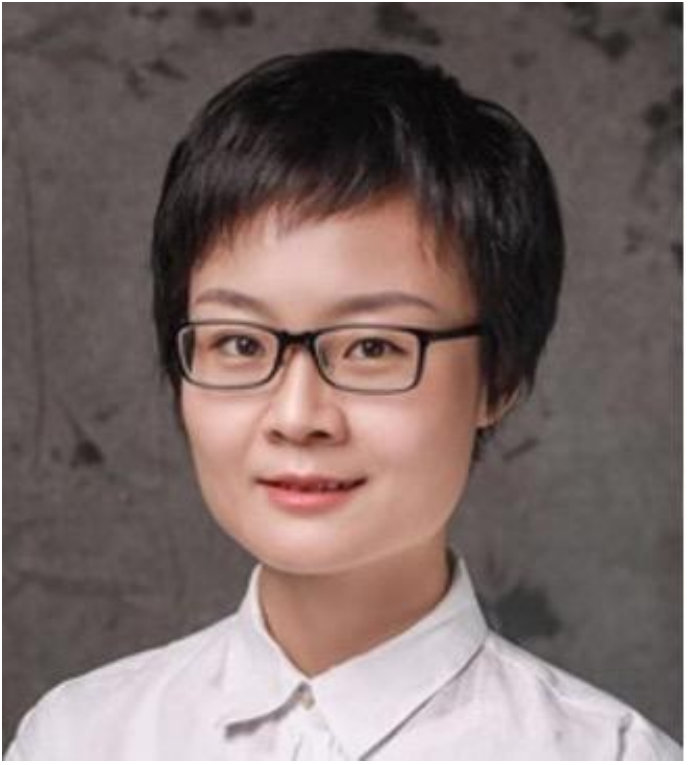}}] {Yang Li}(Member, IEEE) received her B.A. degree in mathematics and computer science from Smith College in 2011, and her Ph.D. degree in computer science from Stanford University in 2017.  She has joined Tsinghua Shenzhen International Graduate School since 2017, currently working as an Associate Professor in the Institute of Data and Information, and a principal investigator in the Shenzhen Key Laboratory of Ubiquitous Data Enabling. Her research focuses on developing trustworthy machine learning methods, particularly in transfer learning, model adaptability, and explainability, with applications in medical image understanding. She serves as an associate editor for Franklin Open and is a member of the editorial board for Digital Signal Processing.
\end{IEEEbiography}


\begin{IEEEbiography}[{\includegraphics[width=1in,height=1.25in,clip,keepaspectratio]{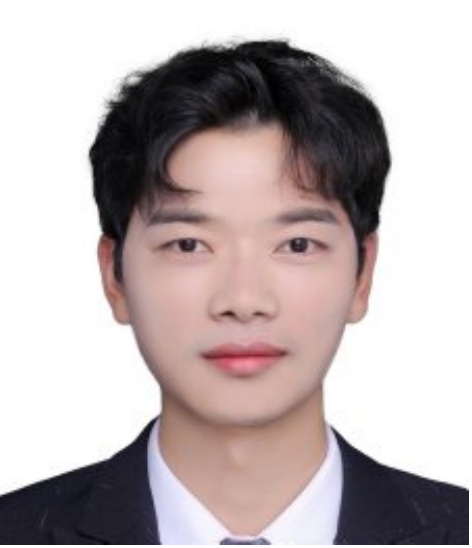}}] {Kui Jiang}(Member, IEEE) received the M.E. and Ph.D. degrees from the School of Computer Science, Wuhan University, Wuhan, China, in 2019 and 2022, respectively. Before July 2023, he was a Research Scientist with the Cloud BU, Huawei. He is currently
an Associate Professor with the School of Computer Science and Technology, Harbin Institute of Technology. He received the 2022 ACM Wuhan Doctoral Dissertation Award. His research interests include image/video processing and computer vision.
\end{IEEEbiography}


\begin{IEEEbiography}[{\includegraphics[width=1in,height=1.25in,clip,keepaspectratio]{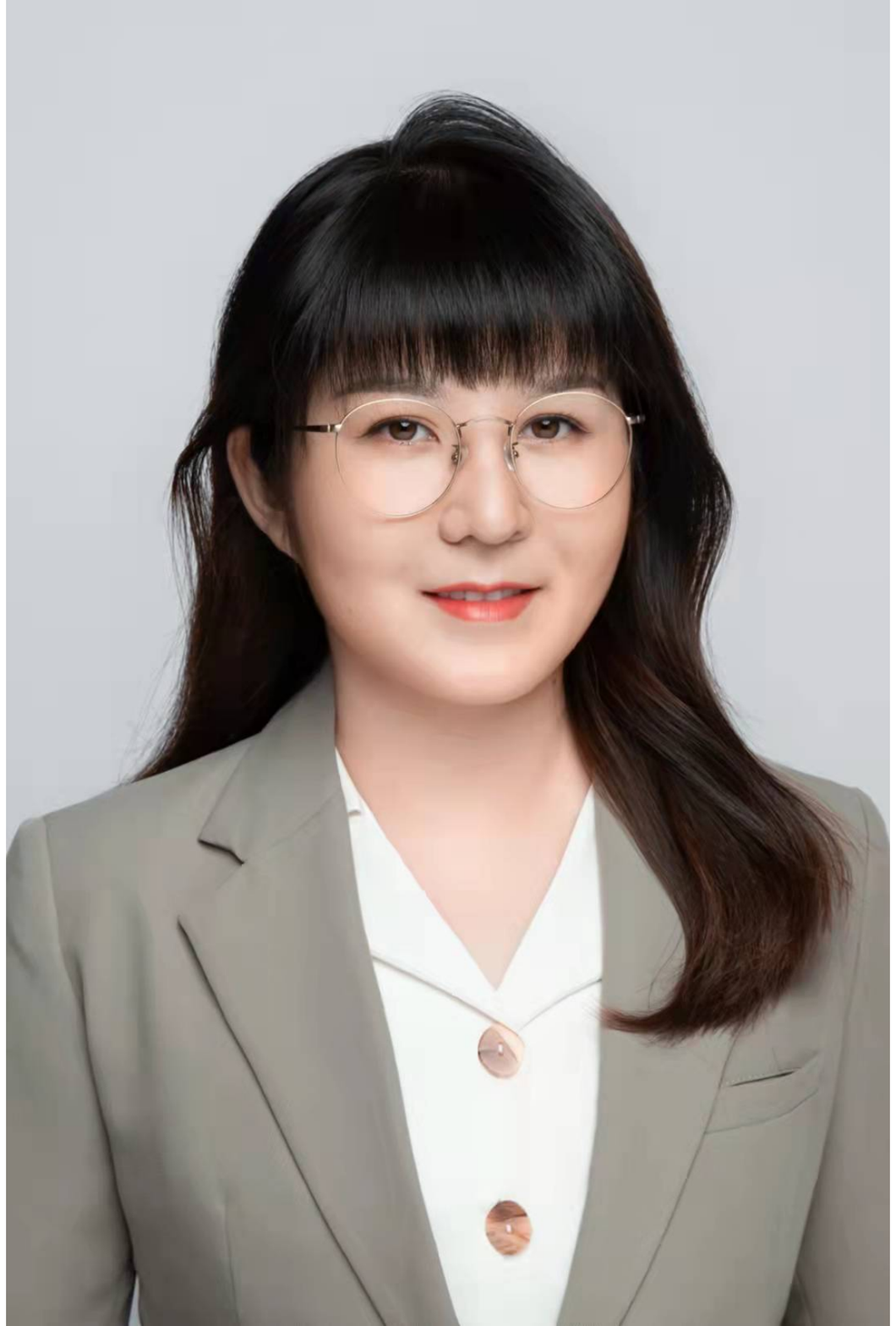}}] {Zihan Geng}(Member, IEEE) is currently an Assistant Professor and a Ph.D. supervisor with the Tsinghua Shenzhen International Graduate School, Tsinghua University, a member of the Optical Society of America (OSA), and is recognized as a high-level overseas talent in Shenzhen. She has published over forty papers in journals and conferences, such as Light: Science and Applications, Photonics Research, and Laser \& Photonics Reviews. She has led many national and provincial-level projects as a principal investigator. From 2010 to 2014, she studied with the School of Information Science and Engineering, Central South University, and Monash University in electronic and computer systems engineering, earning dual undergraduate degrees.  In 2018, she received her Ph.D. degree from Monash University. In 2019, she took charge of the ultra-wideband metropolitan communication project and the multimode transmission project with Huawei’s Central Research Institute, focusing on cutting-edge technology research for the next ten to 20 years. From 2019 to 2020, she was the Head of Strategic Planning with Huawei’s AI Laboratory, she analyzed the latest developments in computer vision, speech semantics, decision reasoning, foundational AI theories, and platforms, aligning cutting-edge AI technology with business scenarios, concentrated on the third generation of AI technology aimed at interpretable, robust, and adaptive AI. In 2020 and 2021, she was an Assistant Professor at the Harbin Institute of Technology. From 2021 to now she has worked at Tsinghua University.
\end{IEEEbiography}

\begin{IEEEbiography}[{\includegraphics[width=1in,height=1.25in,clip,keepaspectratio]{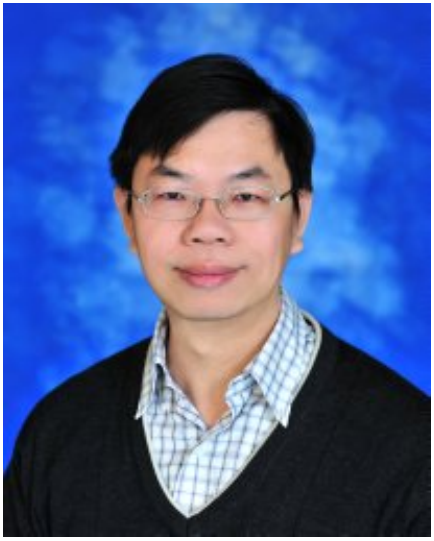}}] {Chia-Wen Lin}
 (Fellow, IEEE) received his Ph.D. degree in electrical engineering from National Tsing Hua University (NTHU), Hsinchu, Taiwan, in 2000. He was with the Department of Computer Science and Information Engineering, National Chung Cheng University, Taiwan, from 2000 to 2007. Prior to joining academia, he was with the Information and Communications Research Laboratories, Industrial Technology Research Institute, Hsinchu, from 1992 to 2000. He is currently a Distinguished Professor with
the Department of Electrical Engineering and the Institute of Communications Engineering, NTHU, where he is also the Deputy Director of the AI Research Center. His research interests lie in image and video processing, computer vision, and video networking. He served as a Distinguished Lecturer for IEEE Circuits and Systems Society from 2018 to 2019, a Steering Committee Member for IEEE TRANSACTIONS ON MULTIMEDIA from 2014 to 2015, and the Chair of the Multimedia Systems and Applications Technical Committee for the IEEE Circuits and Systems Society from 2013 to 2015. His articles received the Best Paper Award of IEEE VCIP 2015, top 10\% paper awards of IEEE MMSP 2013, and the Young Investigator Award of VCIP 2005. He also received the Young Investigator Award presented by the Ministry of Science and Technology, Taiwan, in 2006. He has served as the President of the Chinese Image Processing and Pattern Recognition Association, Taiwan, from 2010 to 2019, the Technical Program Co-Chair of IEEE ICME 2010, the General Co-Chair of IEEE VCIP 2018, and the Technical Program Co-Chair of IEEE ICIP 2019. He has served as an Associate Editor for IEEE TRANSACTIONS ON IMAGE PROCESSING, IEEE TRANSACTIONS ON CIRCUITS AND SYSTEMS FOR VIDEO TECHNOLOGY, IEEE TRANSACTIONS ON MULTIMEDIA, IEEE MULTIMEDIA, and Journal of Visual Communication and Image Representation.
\end{IEEEbiography}

\end{document}